\documentclass{llncs}
\usepackage{amssymb, amsmath, epsfig, subfigure}
\usepackage{sidecap}
\usepackage{xcolor}
\usepackage{dsfont}
\usepackage{stmaryrd}
\usepackage{color}
\DeclareMathAlphabet{\mathonebb}{U}{bbold}{m}{n}

\definecolor{darkGreen}{rgb}{0.01,0.5,0.01}
\newcommand{\one}{\ensuremath{\mathonebb{1}}}

\newcommand{\avirer}[1]{ }

\newcommand{\tobs}{t_{obs}}

\newcommand{\iopl}{{I}^{opl}}
\newcommand{\igang}{{I}^{g}}
\newcommand{\ioplb}{{\bar I}^{opl}}
\newcommand{\kij}{{k i j}}

\newcommand{\Vbkij}{{V^b_{k i j}}}

\newcommand{\bioinspired}{bio-inspired\ }

\title{A  \bioinspired image coder with temporal scalability}
\author{Khaled Masmoudi \and Marc Antonini\inst{1} \and Pierre Kornprobst\inst{2}}
\authorrunning{Khaled Masmoudi et al.} 
\institute{I3S laboratory--UNS--CNRS\\
Sophia-Antipolis, France\\
\email{kmasmoud@i3s.unice.fr},\\
\url{http://www.i3s.unice.fr/~kmasmoud}
\and
NeuroMathComp Team Project--INRIA\\
Sophia-Antipolis, France}
\begin{document}
\topmargin -4pt
\hoffset 1.5mm
\maketitle
\begin{abstract}
We present a novel \bioinspired and dynamic coding scheme for static images.
Our coder aims at reproducing the main steps of the visual stimulus processing in the mammalian retina taking into account its time behavior. The main novelty of this work is to show how to exploit the time behavior of the retina cells to ensure, in a simple way, scalability and bit allocation.
To do so, our main source of inspiration will be the biologically plausible retina model called {\itshape Virtual Retina}. 
Following a similar structure, our model has two stages.
The first stage is an image transform which is performed by the outer layers in the retina.
Here it is modelled by filtering the image with a bank of difference of Gaussians with time-delays. 
The second stage is a time-dependent analog-to-digital conversion which is performed by the inner layers in the retina.
Thanks to its conception, our coder enables scalability and bit allocation across time. Also, 
our decoded images do not show annoying artefacts such as ringing and block effects.
As a whole, this article shows how to capture the main properties of a biological system, here the retina, in order to design a new efficient coder. 
\end{abstract}
\section{Introduction}
\label{sec:intro}
Intensive efforts have been made during the past two decades for the design of lossy image coders yielding several standards such as JPEG and JPEG2000~\cite{Antonini92,Christopoulos00}. 
These compression algorithms, mostly, followed the same conception schema, though, improving considerably the performances in terms of cost and quality. Yet, it became clear now that little is still to be gained if no shift is made in the philosophy underlying the design of coders.

In this paper, we propose a novel image codec based on visual system properties: Our aim is to set a new framework for coder design. In this context, neurophysiologic studies 
``have demonstrated that our sensory systems are remarkably efficient at coding the sensory environment''~\cite{graham2007efficient}, and we are convinced that an interdisciplinary approach would improve coding algorithms. 

We focused on the complex computations that the mammalian retina operates to transform the incoming light stimulus into a set of uniformly-shaped impulses, also called spikes. 
Indeed, recent studies such as \cite{gollisch2010eye} confirmed that the retina is doing non-trivial operations to the input signal before transmission, so that our goal here \textcolor{black}{is} to capture the main properties of the retina processing for the design of our new coder.

Several efforts in the literature reproduced fragments of this retina processing through \bioinspired models and for various vision tasks, for example: object detection and robot movement decision~\cite{linares2007using}, fast categorization~\cite{thorpe1998rank,VanRullen01}, and regions of interest detection for bit allocation~\cite{ouerhani2002adaptive}. 
But most of these approaches do not account for the precise retina processing. Besides, these models overlooked the signal recovery problem which is crucial in the coding application. Attempts in this direction were done making heavy simplifications at the expense of biological relevance~\cite{perrinet2008sparse} or restricting the decoding ability within a set of signals in a dictionary~\cite{pillow2008spatio}. 
Here, the originality of our work  is twofold: we focus explicitly on the coding application and we keep our design as close as possible to biological reality considering most of the mammalian retina processing features.

Our main source of inspiration will be the biologically plausible {\it Virtual Retina} model~\cite{WohrerKornprobst08} whose goal was to find the best compromise between the biological reality and the possibility to make large-scale simulations.
Based on this model, we propose a coding scheme following the architecture and functionalities of the retina, doing some adaptations due to the application.

This paper is organized as follows. In Section~\ref{sec:BiologicalBackground} we revisit the retina model called \textit{Virtual Retina}~\cite{WohrerKornprobst08}. In Section~\ref{sec:coding-pathway}, we show how this retina model can be used as the basis of a novel \bioinspired image coder. The coding pathway is presented in a classical way distinguishing two stages: the image transform and the analog-to-digital (A/D) converter.
In Section~\ref{sec:decoding-pathway} we present the decoding pathway. In  Section~\ref{sec:results} we show the main results that demonstrate the properties of our model. In Section~\ref{sec:discussion} we summarize our main conclusions.

\section{\textit{Virtual Retina}: a bio-inspired retina model}
\label{sec:BiologicalBackground}

The motivation of our work is to investigate the retina functional architecture and use it as a design basis to devise  new codecs. So, it is essential to understand what are the main functional principles of the retina processing. The literature in computational neuroscience dealing with the retina 
proposes different models. 
These models are very numerous, ranking from detailed models of a specific physiological phenomenon, to large-scale models of the whole retina. 

In this article, we focused on the category of large-scale retina models as we are interested in a model that gathers the main features of mammalian retina.
Within this category, we considered the retina model called \textit{Virtual Retina}~\cite{WohrerKornprobst08}. This model is one of the most complete ones in the literature, 
as it encompasses the major features of the actual mammalian retina. This model is mostly state-of-the-art and the authors confirmed its relevance by reproducing accurately real cell recordings for several experiments.

\begin{figure}[t*]
\centerline{\includegraphics[width=0.7\textwidth]{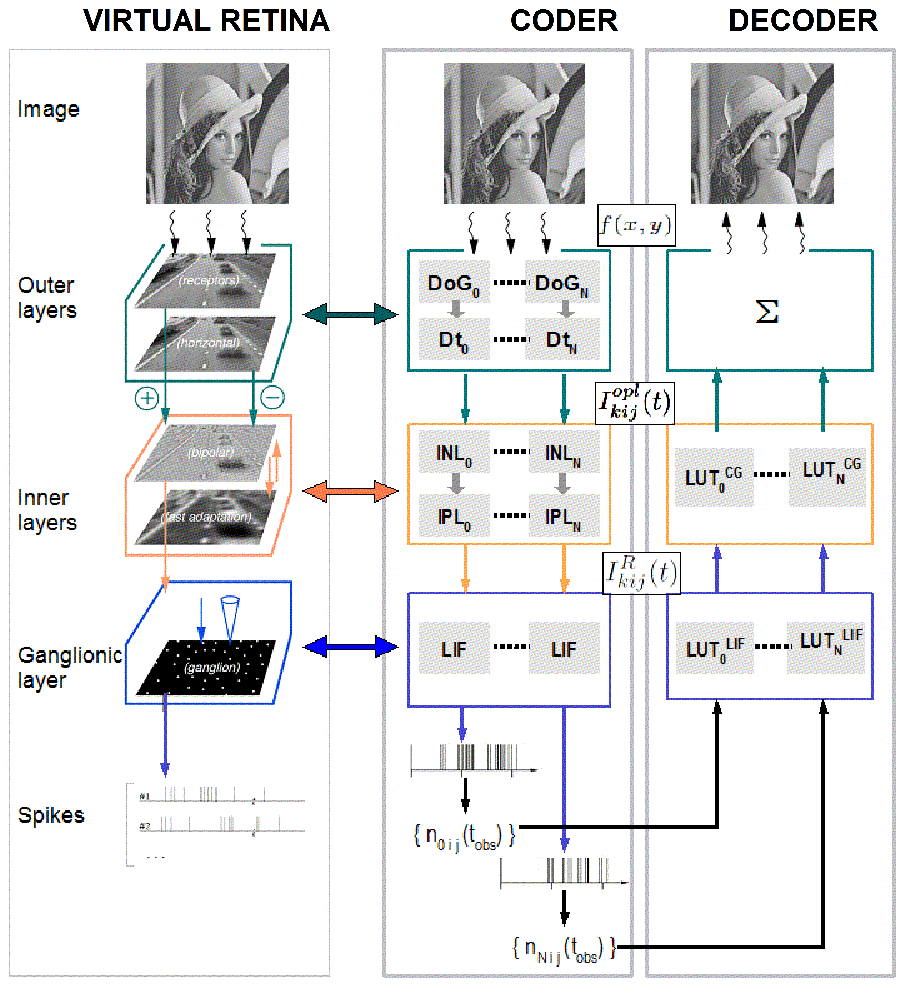}}
 \caption{\label{fig:schema-retina}
(a) Schematic view of the \textit{Virtual Retina} model proposed by \cite{WohrerKornprobst08}.
(b) and (c): Overview of our bio-inspired codec. 
Given an image, the static DoG-based multi-scale transform generates the sub-bands $\left\lbrace F_k \right\rbrace$. DoG filters are sorted from the lowest frequency-band filter $DoG_{0}$ to the highest one $DoG_{N-1}$. Each sub-band $F_k$ is delayed using a time-delay circuit $D_{t_k}$, with $t_k < t_{k+1}$. The time-delayed multi-scale output is then made available to the subsequent coder stages. The final output of the coder is a set of spike series, and the coding feature adopted will be the spike count $n_\kij(t_{obs})$ recorded for each neuron indexed by $(k i j)$ at a given time $t_{obs}$.
}
\end{figure}

The architecture of the \textit{Virtual Retina} model follows the structure of mammalian retina as schematized in Figure~\ref{fig:schema-retina}(a). The model has several interconnected layers 
and three main processing steps can be distinguished:
\begin{itemize}
 \item Outer layers: The first processing step is described by non-separable spatio-temporal filters, behaving as time-dependent edge detectors. This is a classical step implemented in several retina models. 
 \item Inner layers: A non-linear contrast gain control is performed. This step models mainly bipolar cells by control circuits with time-varying conductances.
 \item Ganglionic layer: Leaky integrate and fire neurons are implemented to model the ganglionic layer processing that finally converts the stimulus into spikes.
\end{itemize}


Given this model as a basis, our goal is to adapt it to conceive the new codec presented in the next sections.
\section{The coding pathway}
\label{sec:coding-pathway}
The coding pathway is schematized in Figure~\ref{fig:schema-retina}(b). It follows the same architecture as \textit{Virtual Retina}. However, since we have to define also a decoding pathway, we need to think about the invertibility of each processing stage. For this reason some adaptations are required and described in this section.

\subsection{The image transform: The outer retina layers}
\label{sec:RetinalTransform}
In \textit{Virtual Retina}, the outer layers were modelled by a non-separable spatio-temporal filtering. This processing produces responses corresponding to spatial or temporal variations of the signal because it models time-dependent interactions between two low-pass filters: this is termed center-surround differences. This stage has the property that it responds first to low spatial frequencies and later to higher frequencies. This \textit{time-dependent frequency integration} was shown for \textit{Virtual Retina} (see~\cite{Wohrer09}) and it was confirmed experimentally (see, e.g., \cite{Sterling1992}).
This property is interesting as a large amount of the total signal energy is contained in the lower frequency sub-bands, whereas high frequencies bring further details. This idea already motivated bit allocation algorithms to concentrate the resources for a good recovery on lower frequencies.

However, it appears that inverting this non-separable spatio-temporal filtering is a complex problem~\cite{Wohrer09,zhang2005analytical}. To overcome this difficulty, we propose to model differently this stage while keeping its essential features. 
To do so, we decomposed this process into two steps: The first one considers only center-surround differences in the spatial domain (through differences of Gaussians) which is justified by the fact that our coder here gets static images as input. The second step reproduces the time-dependent frequency integration by the introduction of time-delays.

\subsubsection*{Center-surround differences in the spatial domain: DoG}

Neurophysiologic experiments have shown that, as for classical image coders, the retina encodes the stimulus representation in a transform domain. The retinal stimulus transform is performed in the cells of the outer layers, mainly in the outer plexiform layer (OPL). Quantitative studies such as~\cite{Field94,Rodieck65} have proven that the OPL cells processing can be approximated by a linear filtering. In particular, the authors in~\cite{Field94} proposed the largely adopted DoG filter which is a weighted difference of spatial Gaussians that is defined as follows:
\begin{equation}
\label{eq:dog}
DoG (x, y) = w_c G_{{\sigma}_c}(x,y)-w_s G_{{\sigma}_s}(x,y),
\end{equation}
where $w_c$ and $w_s$ are the respective weights of the center and surround components of the receptive fields, and
 $\sigma_{c}$ and $\sigma_{s}$ are the standard deviations of the Gaussian kernels $G_{{\sigma}_c}$ and $G_{{\sigma}_s}$.

\begin{figure}
\centerline{
\begin{tabular}{c@{\hspace*{2mm}}c@{\hspace*{2mm}}c}
\includegraphics[height=3cm]{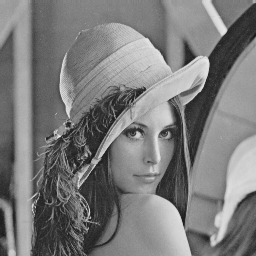}&
\includegraphics[height=3cm]{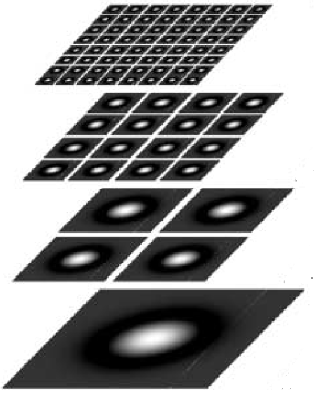}&
\includegraphics[height=3cm]{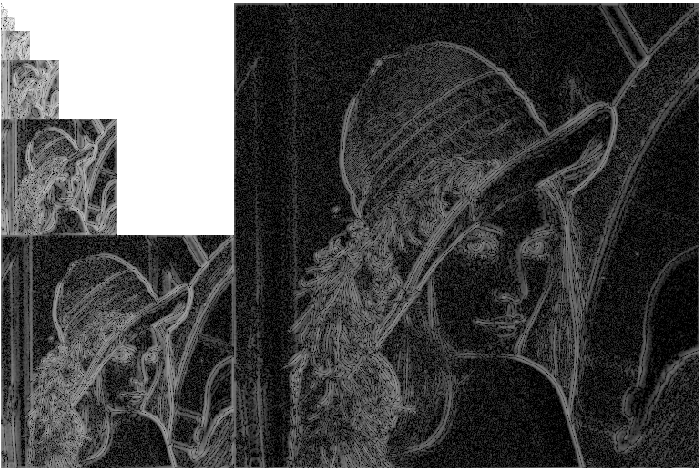} \\
(a)&(b)&(c)
\end{tabular}
}
\caption{\label{fig:dog}
(a) Input Lena Image.
(b) Example of a dyadic grid of DoG's used for the image analysis (from~\cite{VanRullen01}).
(c) Example on image (a) of DoG coefficients generated by the retina model (the sub-bands are shown in the logarithmic scale)
}
\end{figure}

In terms of implementation, as in~\cite{VanRullen01}, the DoG cells can be arranged in a dyadic grid to sweep all the stimulus spectrum as schematized in Figure~\ref{fig:dog}(a). Each layer $k$ in the grid, is tiled with $DoG_k$ cells having a scale $s_k$ 
and generating a transform sub-band ${F}_k$, where $\sigma_{s_{k+1}}=\frac{1}{2}\sigma_{s_k}$. So, in order to measure the degree of activation $\ioplb_\kij$ of a given $DoG_k$ cell at the location $(i, j)$ with a scale $s_k$, we compute the convolution of the original image $f$ by the $DoG_k$ filter:
\begin{equation}
\label{eq:cijs}
\ioplb_\kij = \sum_{x,y=-\infty}^{\infty}DoG_{k}(i-x, j-y) \, f(x, y).
\end{equation}

This generates a set of $\frac{4}{3} N^2-1$ coefficients for an $N^2$-sized image, as it works in the same fashion as a Laplacian pyramid~\cite{Burt1983}. An example of such a \bioinspired multi-scale decomposition 
is shown in Figure~\ref{fig:dog}(b). Note here that we added to this bank of filters a Gaussian low-pass scaling function that represents the state of the OPL filters at the time origin. This yields a low-pass coefficient $\ioplb_{0 0 0}$ and enables the recovery of a low-pass residue at the reconstruction level~\cite{crowley2009fast,Masmoudi2010}. 

\subsubsection*{Integrating time dynamics through time-delay circuits}

Of course, the model described in~\eqref{eq:cijs} has no dynamical properties. In the actual retina, the surround $G_{\sigma_s}$ in~\eqref{eq:dog} appears progressively across time driving the filter passband from low frequencies to higher ones. 
Our goal is to reproduce this phenomenon that we called time-dependent frequency integration. To do so, we added in the coding pathway of each sub-band $F_k$ a time-delay circuit $D_{t_k}$. The value of ${t_k}$ is specific to $F_k$ and is linearly increasing as a function of $k$. The $t_k$-delay causes the sub-band ${F}_k$ to be transmitted to the subsequent stages of the coder starting from the time $t_k$. The time-delayed activation coefficient $I^{opl}_\kij(t)$ computed at the location $(i, j)$ for the scale $s_k$ at time $t$ is now defined as follows:
\begin{equation}
\label{eq:cijskt}
\iopl_\kij(t) = \ioplb_\kij \ \ \one_{ \left\lbrace t\geqslant t_k \right\rbrace }(t),
\end{equation}
where $\one_{ \left\lbrace t\geqslant t_k \right\rbrace }$ is the indicator function such that, $\one_{ \left\lbrace t\geqslant t_k \right\rbrace }(t) = 0$ if $t < t_k$ and $1$ otherwise. 

\avirer{Obviously, actual recordings show that the current is a time-fluctuating function. The step nature of the outer layers output, in our case, is due to the simplified model introduced where the time dimension is recreated artificially. }



\subsection{The A/D converter: inner and ganglionic layers}
\label{sec:ADC}
The retinal A/D converter is defined based on the processing occurring in the inner and ganglionic layers, namely a contrast gain control, a non-linear rectification and a discretization based on leaky integrate and fire (LIF) neurons~\cite{Masmoudi2010another}.
A different treatment will be performed for each delayed sub-band, and this produces a natural bit allocation mechanism. Indeed, as each sub-band $F_k$ is presented at a different time $t_k$, it will be subject to a transform according to the state of our dynamic A/D converter at $t_k$. 

\subsubsection{Contrast gain control}
\label{sec:ContrastGainControl}
Retina adjust its operational range to match the input stimuli magnitude range. This is done by an operation called contrast gain control mainly performed in the bipolar cells. Indeed, real bipolar cells conductance is time varying, resulting in a phenomenon termed \textit{shunting inhibition}. This shunting avoids the system saturation by reducing high magnitudes.\\
In \textit{Virtual Retina}, given the scalar magnitude $\ioplb_\kij$ of the input step current $I^{opl}_\kij (t)$, the contrast gain control is a non-linear operation on the potential of the bipolar cells. This potential varies according to both the time and the magnitude value $\ioplb_\kij$; and will be  denoted by $\Vbkij(t, \ioplb_\kij)$. 

\begin{figure}[ht]
{
\subfigure[\label{fig:vBipTemps21ms}]
{\includegraphics[width=0.48\textwidth, height=0.325\textwidth]{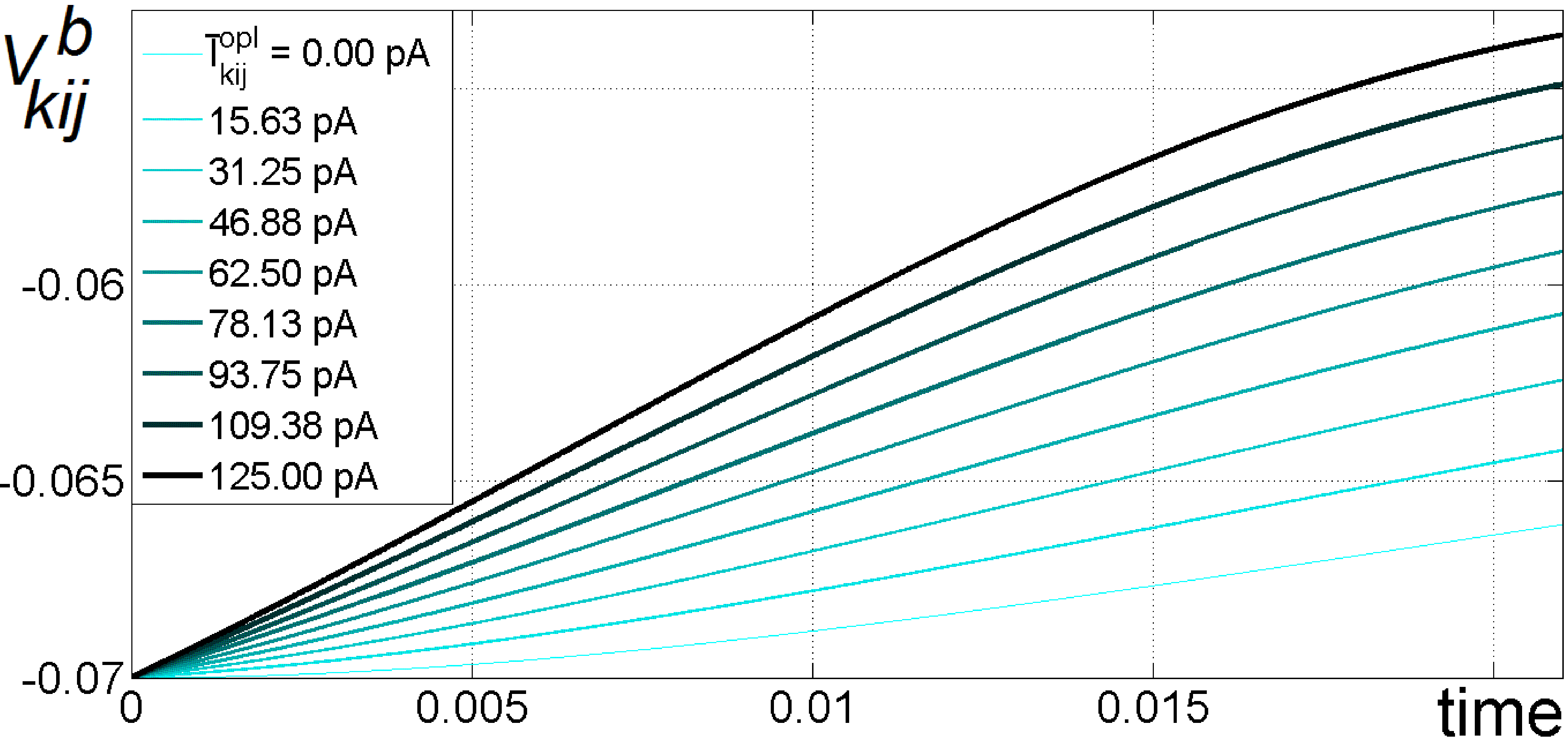}}\hfill
\subfigure[\label{fig:iGangTemps21ms}]
{\includegraphics[width=0.48\textwidth, height=0.325\textwidth]{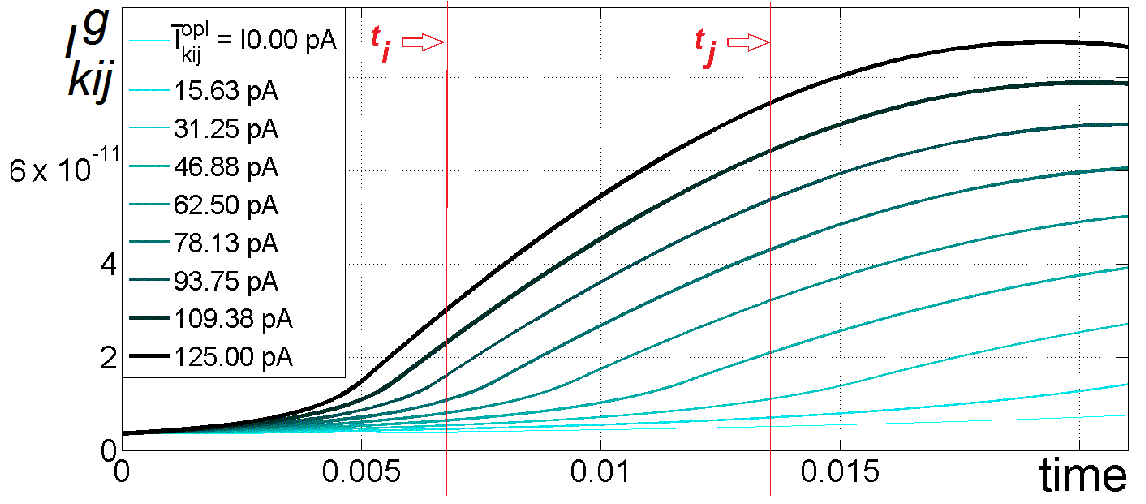}}\\
\subfigure[\label{fig:iGangIoplCoupes}]
{\includegraphics[width=0.48\textwidth, height=0.325\textwidth]{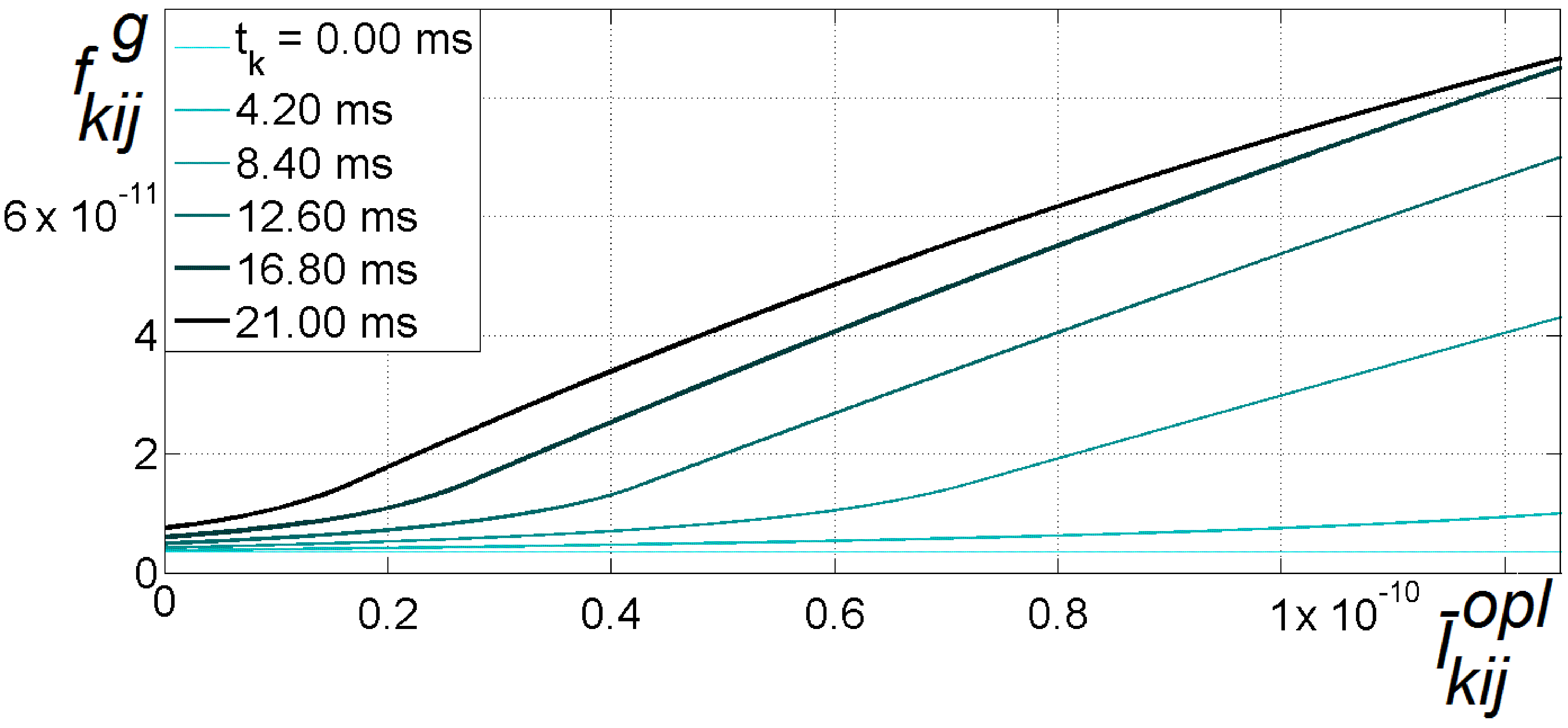}}\hfill
\subfigure[\label{fig:neuroneLifCoupes100ms}]
{\includegraphics[width=0.48\textwidth, height=0.325\textwidth]{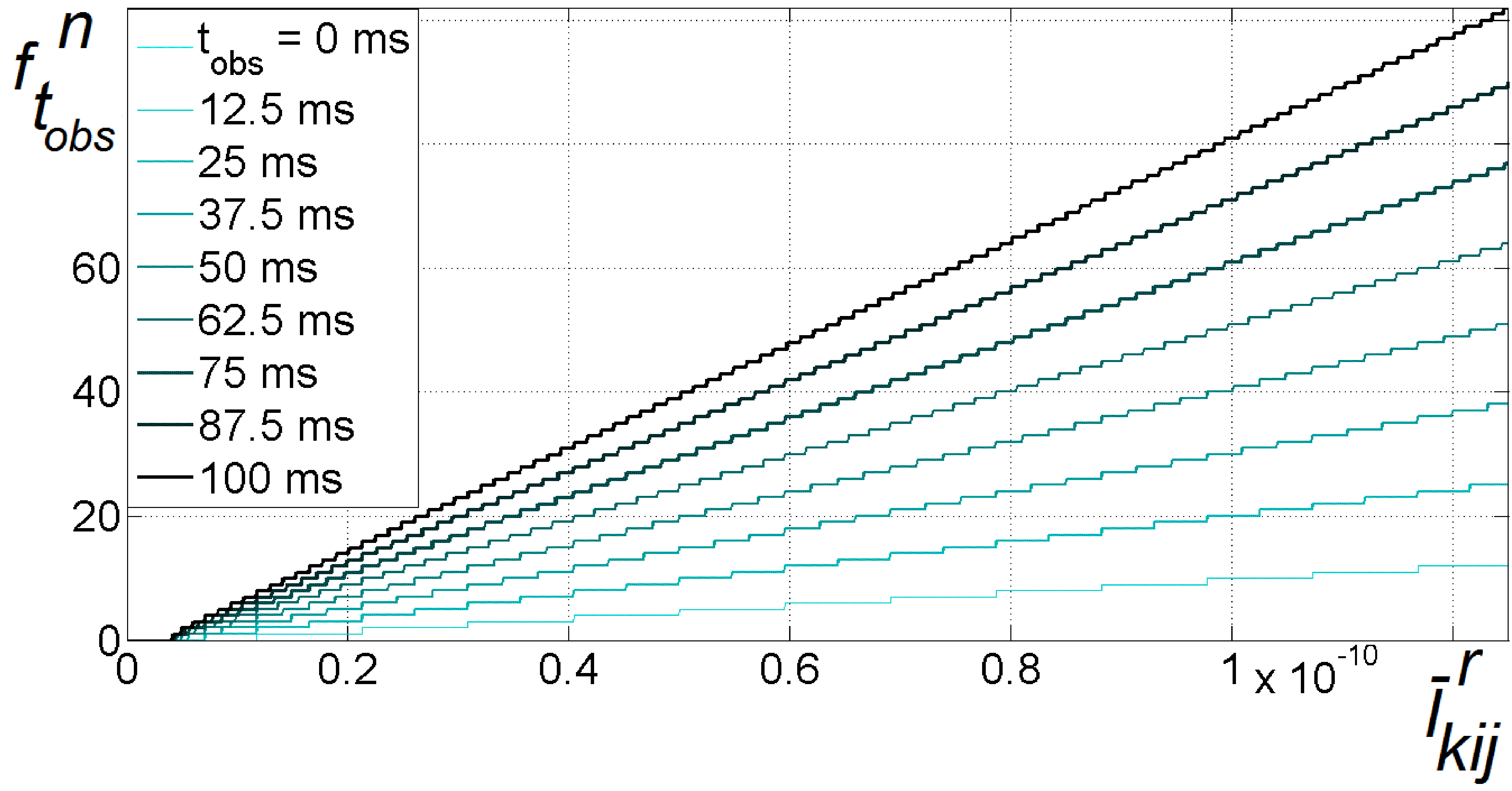}}
}
\caption{\label{fig:inner-layers-action}
\ref{fig:vBipTemps21ms}: $\Vbkij(t)$ as a function of time for different values of $\ioplb$;
\ref{fig:iGangTemps21ms}: $\igang_{\kij}$ as a function of time for different values of $\ioplb$;
\ref{fig:iGangIoplCoupes}: The functions $f^g_{t_k}$ that map $\ioplb_\kij$ into $\igang_{\kij}$ for different values of $t_k$; 
\ref{fig:neuroneLifCoupes100ms}: The functions $f^n_{t_{obs}}$ that map  $\bar I_\kij^r$ into $n_{\kij}$ for different values of $t_{obs}$
}
\end{figure}

This phenomenon is modelled, for a constant value of $\ioplb_\kij$, by the following differential equation: 
\begin{equation}
\label{eq:vBip}
\left\{
\begin{array}{l}
c^b\dfrac{d \Vbkij(t, \ioplb_\kij)}{dt}+g^b(t)\Vbkij ( t, \ioplb_\kij ) = \iopl_\kij (t),\quad \mbox{for } t \geqslant 0,\\
g^b (t) = E_{\tau^b} \stackrel{t}{*} Q (\Vbkij(t, \ioplb_\kij )),
\end{array}
\right.
\end{equation}
where $Q(\Vbkij) = g_0^b + \lambda^{b} \left(\Vbkij(t)\right)^2$ and $E_{\tau^{b}}=\dfrac{1}{\tau^{b}} exp^{\frac{-t}{{\tau}^{b}}}, \mbox{ for }t\geqslant 0$.
Figure~\ref{fig:vBipTemps21ms} shows the  time behavior of $\Vbkij(t, \ioplb_\kij)$ for different  magnitude values $\ioplb_\kij$ of $\iopl_\kij(t)$. 

\subsubsection{Non-linear rectification}

Then, the potential  $\Vbkij(t, \ioplb_\kij)$ is subject to a non-linear rectification yielding the so-called ganglionic current $\igang_\kij (t, \ioplb_\kij )$. {\it Virtual Retina} models it, for a constant scalar value $\ioplb_\kij$,  by: 
\begin{equation}
\label{eq:IPLCorrection}
\igang_\kij(t, \ioplb_\kij) = N\left( T_{w^{g}, \tau^{g}}(t) * \Vbkij(t, \ioplb_\kij)\right),\quad \mbox{for } t \geqslant 0,
\end{equation}
where $w^{g}$ and $\tau^{g}$ are constant scalar parameters, 
$T_{w^{g}, \tau^{g}} $ is the linear transient filter defined by 
$T_{w^{g}, \tau^{g}} = \delta_0(t) - w^{g} E_{\tau^{g}}(t),$
and $N$ is defined by:
$$N(v) = \left\{
\begin{array}{l}
  \dfrac{i^{g}_0}{i^{g}_0-\lambda^{g}(v-v^{g}_0)},\mbox{ if } v < v^{g}_0\\
  i^{g}_0+\lambda^{g}(v-v^{g}_0),\mbox{ if } v \geqslant v^{g}_0,
\end{array}
\right.$$
where $i^g_0$, $v^g_0$, and $\lambda^g$ are constant scalar parameters. Figure~\ref{fig:inner-layers-action}(b) shows the time behavior of $\igang_{\kij}(t, \ioplb_\kij)$ for different values of $\ioplb_\kij$. 

As the currents $\ioplb_\kij$ are delayed with times $\{ t_k \}$, our goal is to catch the instantaneous behavior of the inner layers at these times $\{ t_k \}$. This amounts to infer the transforms $I^{g}_{t_k} (\ioplb_\kij)$ that maps a given scalar magnitude $\ioplb_\kij$ into a rectified current $\bar I^r_\kij$ as the modelled inner layers would generate it at $t_k$. 
To do so, we start from the time-varying curves of $\igang_{\kij}(t, \ioplb_\kij)$ in Figure~\ref{fig:iGangTemps21ms} and we do a transversal cut at each time $t_{k}$: We show in Figure~\ref{fig:iGangIoplCoupes} the resulting maps $f^g_{t_k}$ such that $\igang_{kij}(t_k, \ioplb_\kij) = f^g_{t_k}(\ioplb_\kij)$.

As for $\iopl_\kij(t)$ (see~\eqref{eq:cijskt}), we introduce the time dimension using the indicator function $\one_{ \left\lbrace t\geqslant t_k \right\rbrace }(t)$. The final output of this stage is the set of step functions $I^r_\kij(t)$ defined by:
\begin{align}
I^r_{k i j}(t) =\bar I^r_\kij\; \one_{ \left\lbrace t\geqslant t_k \right\rbrace }(t),\, \mbox{ with }\bar I^r_\kij = f^g_{t_k} (\ioplb_\kij).
\end{align}

This non-linear rectification is analogous to a widely-used telecommunication technique:  the companding~\cite{refCompandor}. Companders are used to make the quantization steps unequal after a linear gain control stage. Though, unlike $A-law$ or $\mu-law$ companders that amplify low magnitudes, the inner layers emphasize high magnitudes in the signal. 
Besides, the inner layers stage have a time dependent behavior, whereas a usual gain controller/compander is static, and this makes our A/D converter go beyond the standards.

\subsubsection{\label{sec:quantization}Leaky integrate-and-fire quantization:}
The ganglionic layer is the deepest one tiling the retina: it transforms a continuous signal $I^r_\kij(t)$ into discrete sets of spike trains. As in {\it Virtual Retina}, this stage is modelled by leaky integrate and fire neurons (LIF) which is a classical model. One LIF neuron is associated to every position in each sub-band $F_k$. The time-behavior of a LIF neuron is governed by the fluctuation of its voltage $V_\kij(t)$. Whenever $V_\kij(t)$ reaches a predefined $\delta$ threshold, a spike is emitted and the voltage goes back to a resting potential $V^0_R$. 
Between two spike emission times, $t_\kij^{(l)}$ and $t_\kij^{(l+1)}$, the potential evolves according to the following differential equation:
\begin{equation}
\label{eq:voltageOutputSpikingNeuron}
c^l\dfrac{dV_\kij(t)}{dt} + g^l V_\kij(t) = I^r_\kij (t),\quad \mbox{ for } t \in [ t_\kij^{(l)},\, t_\kij^{(l+1)}],
\end{equation}
where $g^l$ is a constant conductance, and $c^l$ is a constant capacitance. 
In the literature, neurons activity is commonly characterized by the count of spikes emitted during an observation time bin $[0, t_{obs}]$, which we denote by $n_{k i j}(t_{obs})$~\cite{Gerstner02}. Obviously, as $n_{k i j}(t_{obs})$ encodes for the value of $I^r_\kij (t)$, there is a loss of information as  $n_{k i j}(t_{obs})$  is an integer. The LIF is thus performing a quantization. 
If we observe the instantaneous behavior of the ganglionic layer at different times $t_{obs}$, we get a quasi-uniform scalar quantizer that refines in time. We can do this by a similar process to the one described in the previous paragraph.
We show in Figure~\ref{fig:neuroneLifCoupes100ms} the resulting maps $f^n_{t_{obs}}$ such that $n_{kij}(t_{obs}) = f^n_{t_{obs}}(\bar I^r_\kij)$.


Based on the set $\{n_{k i j}(t_{obs})\}$, measured at the output of our coder, we describe in the next section the decoding pathway to recover the initial image $f(x,y)$. 



\section{The decoding pathway}
\label{sec:decoding-pathway}
The decoding pathway is schematized in Figure~\ref{fig:schema-retina}(c). It consists in inverting, step by step, each coding stage described in Section~\ref{sec:coding-pathway}. At a given time $t_{obs}$, the coding data is the set of $(\frac{4}{3}N^2-1)$ spike counts $n_\kij(t_{obs})$, this section describes how we can recover an estimation $\tilde{f}_{t_{obs}}$ of the $N^2$-sized input image $f(x,y)$. 
Naturally, the recovered image $\tilde{f}_{t_{obs}}(x, y)$ depends on the time $t_{obs}$ which ensures time-scalability: the quality of the reconstruction improves as $t_{obs}$ increases.
The ganglionic and inner layers are inverted using look-up tables constructed off-line and the image is finally recovered by a direct reverse transform of the outer layers processing. 

\subsubsection*{Recovering the input of the ganglionic layer:} 
First, given a spike count $n_\kij(\tobs)$, we recover $\tilde{I}^{r}_\kij(t_{obs})$, the estimation of ${I}^{r}_{k i j}(t_{obs})$. 
To do so, we compute off-line the look-up table $n_{t_{obs}}(\bar I^r_\kij)$ that maps the set of current magnitude values $\bar I^r_\kij$ into spike counts at a given observation time $t_{obs}$ (see Figure~\ref{fig:neuroneLifCoupes100ms}). The reverse mapping is done by a simple interpolation in the reverse-look up table denoted $LUT^{LIF}_{t_{obs}}$. 
Here we draw the reader's attention to the fact that, as the input of the ganglionic layer is delayed, each coefficient of the sub-band $F_k$ is decoded according to the reverse map $LUT^{LIF}_{t_{obs}-t_k}$. Obviously, the recovered coefficients  do not match exactly the original ones due to the quantization performed in the LIF's. 

\subsubsection*{Recovering the input of the inner layers:}

Second, given a rectified current value $\tilde{I}^{r}_\kij(t_{obs})$, we recover $\tilde{I}^{opl}_\kij(t_{obs})$, the estimation of $\iopl_\kij(t_{obs})$. 
In the same way as for the preceding stage, we infer the reverse ``inner layers mapping''  through the pre-computed look up table $LUT_{t_{obs}}^{CG}$ . The current intensities $\tilde{I}^{opl}_\kij(t_{obs})$, corresponding to the retinal transform coefficients, are passed to the subsequent retinal transform decoder. 

\subsubsection*{Recovering the input stimulus:} 
Finally, given the set of $\frac{4}{3}N^2-1$ coefficients $\{\tilde{I}^{opl}_\kij(t_{obs})\}$, we recover ${\tilde{f}_{t_{obs}}}(x, y)$, the estimation of the original image stimulus $f(x, y)$. 
Though the dot product of every pair of $DoG$ filters is approximately equal to $0$, the set of filters considered is not strictly orthonormal. We proved in~\cite{MasmoudiTNN2011} that there exists a dual set of vectors 
enabling an exact reconstruction. 
Hence, the reconstruction estimate ${\tilde{f}}$ of the original input $f$ can be obtained as follows:
\begin{equation}
\label{eq:definitionReconstruction}
{\tilde{f}_{t_{obs}}}(x, y) = 	\sum_{ \left\lbrace\kij\right\rbrace }
					 \tilde{I}^{opl}_{\kij}(t_{obs}) \, \widetilde{DoG}_{k}(i-x, j-y),
\end{equation}
where ${ \left\lbrace\kij\right\rbrace }$ is the set of possible scales and locations in the considered dyadic grid and $\widetilde{DoG}_{k}$ are the duals of the  ${DoG}_{k}$ filters obtained as detailed in~\cite{MasmoudiTNN2011}.
Equation~\eqref{eq:definitionReconstruction} defines a  progressive reconstruction depending on $t_{obs}$. This feature makes the coder be time-scalable.

\section{Results}
\label{sec:results}

We show examples of image reconstruction using our bio-inspired coder at different times\footnote{In all experiments, the model parameters are set to biologically realistic values:
$g^b_0 = 8 \, 10^-{10}\, S$,  
$\tau^b = 12 \, 10^{-3}\, s$, 
$\lambda^b= 9 \, 10^{-7}$,  
$c^b = 1.5 \, 10^{-10}\,F$, 
$v^{g}_0 = 4 \, 10^{-3}\, V$,    
$i^{g}_0 = 15 \, 10^-12\, A$,    
$w^{g}  = 8 \, 10^-1$,     
$\tau^{g} = 16 \, 10^{-3}\, s$;    
$\lambda^{g} = 12 \, 10^{-9}\, S$,    
$\delta = 2\, 10^{-3}\, V,$         
$g^L = 2\,10^{-9}\, S,$          
$V^0_R = 0\, V$, 
$t_k = 5\,10^{-3}\,+\,k\,10^{-3}s$.                 
}.  
Then, we study these results in terms of quality and bit-cost. 
\\
Quality is assessed by classical image quality criteria (PSNR and mean SSIM~\cite{Wang04}). The cost is measured by the Shannon entropy $H(t_{obs})$ upon the population of $\{n_\kij(t_{obs})\}$. The entropy computed in bits per pixel ($bpp$), for an $N^2$-sized image, is defined by:
$
H(t_{obs}) = \frac{1}{N^2}\sum_{k=0}^{K-1} 
2^{2k} H\left(\left\lbrace n_{s_k i j}(t_{obs}), (i,j)\in {\llbracket 0, 2^k-1 \rrbracket}^2 
\right\rbrace\right) 
$,\\
 where $K$ is the number of analyzing sub-bands.

\begin{figure}
\centerline{
\begin{tabular}{c@{\hspace*{1mm}}c@{\hspace*{1mm}}c@{\hspace*{1mm}}c}
\includegraphics[width=0.24\textwidth]{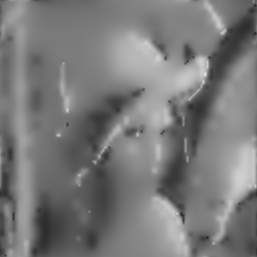}&
\includegraphics[width=0.24\textwidth]{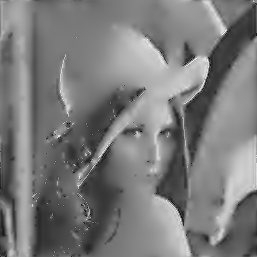}&
\includegraphics[width=0.24\textwidth]{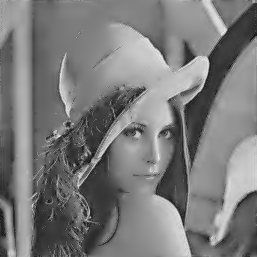}&
\includegraphics[width=0.24\textwidth]{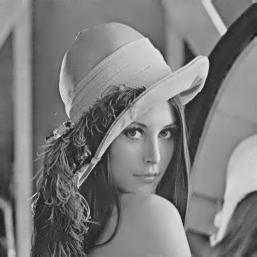}\\
\includegraphics[width=0.24\textwidth]{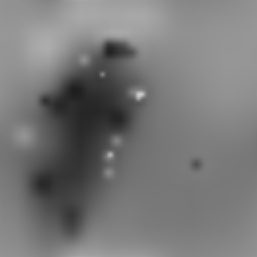}&
\includegraphics[width=0.24\textwidth]{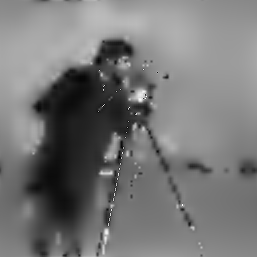}&
\includegraphics[width=0.24\textwidth]{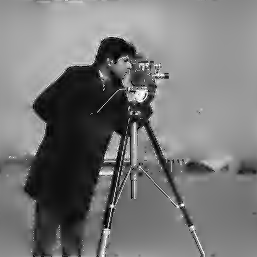}&
\includegraphics[width=0.24\textwidth]{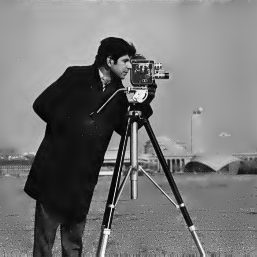}
\end{tabular}
}
\caption{\label{fig:reconstruction}Progressive image reconstruction of Lena and Cameraman using our new bio-inspired coder. The coded/decoded image is shown at: 20 ms, 30 ms, 40 ms, and 50 ms. Rate/Quality are computed for each image in terms of the triplet (bit-cost in $bpp$/ PSNR quality in $dB$/ mean SSIM quality). Upper line: From left to right (0.07 $bpp$/ 20.5 $dB$/ 0.59), (0.38 $bpp$/ 24.4 $dB$/ 0.73), (1.0 $bpp$/ 29.1 $dB$/ 0.86), and (2.1 $bpp$/ 36.3 $dB$/ 0.95).
Lower line: From left to right (0.005 $bpp$/ 15.6 $dB$/ 0.47), (0.07 $bpp$/ 18.9 $dB$/ 0.57), (0.4 $bpp$/ 23 $dB$/ 0.71), and (1.2 $bpp$/ 29.8 $dB$/ 0.88).
}
\end{figure}

Figure~\ref{fig:reconstruction} shows two examples of progressive reconstruction obtained with our new coder. The new concept of \textit{time scalability} is an interesting feature as it introduces time dynamics in the design of the coder. This is a consequence of the mimicking of the actual retina. 
We also notice that, as expected, low frequencies are transmitted first to get a first approximation of the image, then details are added progressively to draw its contours. 
The bit-cost of the coded image is slightly high. This can be explained by the fact that Shannon entropy is not the most relevant metric in our case as no context is taken into consideration, especially the temporal context. Indeed, one can easily predict the number of spikes at a given time $t$ knowing $n_\kij(t-dt)$. 
Note also that no compression techniques, such that bit-plane coding, are yet employed. Our paper aims mainly at setting the basis of new \bioinspired coding designs. 

For the reasons cited above, the performance of our coding scheme in terms of bit-cost have still to be improved to be competitive with the well established JPEG and JPEG2000 standards. Thus we show no comparison in this paper. Though primary results are encouraging, noting that optimizing the bit-allocation mechanism and exploiting coding techniques as bit-plane coding~\cite{taubman2000high} would improve considerably the bit-cost. Besides, the image as reconstructed with our bio-inspired coder shows no ringing and no block effect. 
Finally our codec enables scalability in an original fashion through the introduction of time dynamics within the coding mechanism.

Note also that differentiation in the processing of sub-bands, introduced through time-delays in the retinal transform, enables implicit but still not optimized bit-allocation. In particular the non-linearity in the inner layers stage amplifies singularities and contours, and these provide crucial information for the analysis of the image. The trade-off between the emphasize made on high frequencies and the time-delay in the starting of their coding process is still an issue to investigate.

\section{Conclusion}
\label{sec:discussion}
We proposed a new bio-inspired codec for static images. 
The image coder is based on two stages. The first stage is the image transform as performed by  the outer layers of the retina. In order to integrate time dynamics, we added to this transform time delays that are sub-band specific so that, each sub-band is processed differently. The second stage is a succession of two dynamic processing steps mimicking the deep retina layers behavior. The latter perform an A/D conversion and generate a spike-based, invertible, retinal code for the input image in an original fashion.


Our coding scheme offers interesting features such as (i) time-scalability, as the choice of the observation time of our codec enables different reconstruction qualities, and (ii) bit-allocation, as each sub-band of the image transform is separately mapped according to the corresponding state of the inner layers. 
Primary results are encouraging, noting that optimizing the bit-allocation  and using coding techniques as bit-plane coding would improve considerably the cost.
\avirer{Results in terms of coding cost, though still inferior, are to be considered with nuances.  First, Shannon entropy is not the most relevant metric in our case as no context is analyzed, especially the temporal context. In addition, further optimization of the bit-allocation mechanism as well as encoding techniques, such that bit-plane coding, are to be employed. }
 

This work is at the crossroads of diverse hot topics in the fields of neurosciences, brain-machine interfaces, and signal processing and tries to lay the groundwork for future efforts, especially concerning the design of new biologically inspired coders.


\bibliographystyle{splncs03}

\end{document}